\documentclass[11pt]{article}
\usepackage{acl2016}
\usepackage{times}
\usepackage{latexsym}
\usepackage{graphicx}
\usepackage{fancyvrb}
\usepackage{booktabs}
\usepackage[round]{natbib}
\usepackage[hyphens]{url}
\usepackage{paralist}
\usepackage[table,xcdraw]{xcolor}
\usepackage{amsmath}
\usepackage{dblfloatfix}
\usepackage{pgfplots}
\usepackage{multirow}
\usepackage{color}
\usepackage{amssymb}

\usetikzlibrary{patterns}

\newcommand{\lone}{\text{L1}}

\newcommand{\corpus}{C}
\newcommand{\native}{\corpus_\textrm{N}}
\newcommand{\nonnative}{\corpus_\textrm{NN}}
\newcommand{\translated}{\corpus_\textrm{T}}
\newcommand{\corpusall}{\corpus_\textrm{ALL}}
\newcommand{\fmetric}{f^m}
\newcommand{\fmetriccor}{\fmetric(\corpus)}
\newcommand{\distotal}{D_\textrm{total}}
\newcommand{\disnntn}{D_\textrm{dif}}
\newcommand{\textnl}{\textsl}

\defaultleftmargin{0em}{}{}{}

\aclfinalcopy 

\title{On the Similarities Between \\ Native, Non-native and Translated Texts}

\author{
\fontsize{11}{10}\selectfont{
\begin{tabular}[t]{c@{\extracolsep{4em}}ccc}
Ella Rabinovich$^{\vartriangle\star}$ & Sergiu Nisioi$^{\diamond}$ & Noam Ordan$^{\dagger}$ & Shuly Wintner$^{\star}$ \\
\end{tabular}
}
\\
\fontsize{11}{10}\selectfont{$^{\vartriangle}$IBM Haifa Research Labs} \\
\fontsize{11}{10}\selectfont{$^{\star}$Department of Computer Science, University of Haifa} \\
\fontsize{11}{10}\selectfont{$^{\diamond}$Solomon Marcus Center for Computational Linguistics, University of Bucharest} \\
\fontsize{11}{10}\selectfont{$^{\dagger}$The Arab College for Education, Haifa} \\
\fontsize{10.5}{10}\selectfont{\tt \{ellarabi,sergiu.nisioi,noam.ordan\}@gmail.com, \tt shuly@cs.haifa.ac.il} \\
}

\date{}

\begin{document}

\maketitle

\begin{abstract}
  We present a computational analysis of three language varieties:
  native, advanced non-native, and translation. Our goal is to
  investigate the similarities and differences between non-native
  language productions and translations, contrasting both with native
  language. Using a collection of computational methods we establish
  three main results: (1) the three types of texts are easily
  distinguishable; (2) non-native language and translations are closer
  to each other than each of them is to native language; and (3) some of
  these characteristics depend on the source or native language,
  while others do not, reflecting, perhaps, unified principles that
  similarly affect translations and non-native language.
\end{abstract}

\section{Introduction}
This paper addresses two linguistic phenomena: translation and non-native language. Our main goal is to investigate the similarities and differences between these two phenomena, and contrast them with native language. In particular, we are interested in the reasons for the differences between translations and originals, on one hand, and native and non-native language, on the other. Do they reflect ``universal'' principles, or are they dependent on the source/native language?

Much research in translation studies indicates that translated texts
have unique characteristics. Translated texts (in any language)
constitute a sub-language of the target language, sometimes referred
to as \textit{translationese} \citep{Gellerstam:1986}.
The unique characteristics of translationese have been
traditionally classified into two categories: properties that stem
from \emph{interference} of the source language \citep{Toury:1979},
and universal traits resulting from the translation process itself,
independently of the specific source and target languages
\citep{Baker:1993,Toury:1995}.  The latter so-called \emph{translation
  universals} have triggered a continuous debate among translation studies
  researchers \citep{Universals:2004,house:2008,becher:2010}.

Similarly, over half a century of research on second language
acquisition (SLA) established the presence of \textit{cross-linguistic influences} (CLI)
in non-native utterances \citep{jarvis2008crosslinguistic}.
CLI is a cover term proposed by \citet{kellerman1986crosslinguistic} to denote various
phenomena that stem from language contact situations such as transfer,
interference, avoidance, borrowing, etc.\footnote{To avoid terminological conflicts,
we shall henceforth use CLI to denote any influence of one linguistic system over another,
w.r.t. both translations and non-native productions.}
In addition, universal traits resulting from the learning process itself have been noticed regardless of the native language, \lone.%
\footnote{For simplicity, we will use \emph{\lone} to refer
  both to the native language of a speaker and to the source language
  of a translated text. We use \emph{target language} to refer to
  second and translation languages (English in this paper).}
For example, similar developmental sequences have been observed for negation, question formation,
and other sentence structures in English \citep{dulay1974natural, odlin1989language} for both Chinese and Spanish natives.
Phenomena such as overgeneralization, strategies of learning
\citep{Selinker1972}, psychological factors \citep{ellis1985understanding},
and cultural distance \citep{Giles82} are also influential in the acquisition process.

There are clear similarities between translations and non-native
language: both are affected by the simultaneous presence of (at least) two linguistic systems, which
may result in a higher cognitive load \citep{shlesinger2003effects}. The presence of the L1
may also cause similar CLI effects on the target
language.

On the other hand, there are reasons to believe that translationese and
non-native language should differ from each other.
Translations are produced by
\emph{native} speakers of the target language.
Non-natives, in
contrast, arguably never attain native-like abilities
\citep{coppieters87,johnson1991critical}, however this hypothesis is strongly debated in the SLA community
\citep{birdsong92,lardiere2006ultimate}.

Our goal in this work is to investigate three language \emph{varieties}:
 the language of native speakers (N),
 the language of advanced, highly fluent non-native speakers (NN),
 and translationese (T). We use the term \emph{constrained language} to
refer to the latter two varieties.
We propose a unified computational umbrella for exploring
two related areas of research on bilingualism:
translation studies and second language acquisition.
 Specifically, we put forward three main
hypotheses: %
\begin{inparaenum}[(1)]
\item The three language varieties have unique characteristics that make them
  easily distinguishable.
\item Non-native language and translations are closer to each other
  than either of them is to native language.
\item Some of these characteristics are dependent on the specific
  \lone, but many are not, and may reflect unified principles that
 similarly affect translations and non-native language.
\end{inparaenum}

We test these hypotheses using several corpus-based
computational methods. We use supervised and unsupervised
classification (Section~\ref{sec:classification}) to show that the
three language varieties are easily distinguishable. In particular, we show that
native and advanced non-native productions can be accurately separated. More pertinently, we
demonstrate that non-native utterances and translations comprise two distinct
linguistic systems.

In Section~\ref{sec:similarities}, we use statistical analysis to explore
the unique properties of each language variety. We show that the two varieties of
constrained language are much closer to
each other than they are to native language:
they exhibit poorer lexical richness, a tendency to
use more frequent words, a different distribution of idiomatic
expressions and pronouns, and excessive use of cohesive devices.
This is an unexpected finding, given that both natives and translators (in
contrast to non-natives) produce texts in their mother tongue.

Finally, in Section~\ref{sec:interference} we use language modeling to
show that translations and non-native language exhibit similar
statistical properties that clearly reflect cross-linguistic influences: experiments with
distinct language families reveal salient ties between the two varieties
of constrained language.

The main contribution of this work is thus theoretical:
it sheds light on some fundamental questions regarding
bilingualism, and we expect it to motivate and
drive future research in both SLA and translation
studies. Moreover, a better understanding of constrained language
may also have some practical import, as we briefly
mention in the following section.

\section{Related work}
\label{sec:related-work}

Corpus-based investigation of translationese has been a prolific field
of recent research, laying out an empirical foundation for the
theoretically motivated hypotheses on the characteristics of
translationese. More specifically, identification of translated texts
by means of automatic classification shed light on the manifestation
of translation universals and cross-linguistic influences as markers of translated texts
\citep{Baroni2006, Halteren08, gaspariBernardini, Kurokawa:etal:2009,
  koppel-ordan:2011:ACL-HLT2011, Ilisei:2012, vered:noam:shuly,
  TACL618, nisioi2015b}, while \cite{gaspariBernardini} introduced a dataset for investigation
  of potential common traits between translations and non-native texts.
Such studies prove to be important for
the development of parallel corpora \citep{resnik:smith:03}, the improvement
in quality of plagiarism detection \citep{potthast2011cross}, language modeling,
and statistical machine translation \citep{lembersky-ordan-wintner:CL2012,lembersky-ordan-wintner:CL2014}.

Computational approaches also proved
beneficial for theoretical research in second language
acquisition \citep{jarvis2008crosslinguistic}.
Numerous studies address linguistic processes attributed
to SLA, including automatic detection of highly competent non-native
writers \citep{tomokiyo2001you, bergsma2012stylometric},
identification of the mother tongue of English learners \citep{koppel2005determining,
  tetreault-blanchard-cahill:2013:BEA, tsvetkov-EtAl:2013:BEA8,
  nisioi2015a} and typology-driven error prediction in learners'
speech \citep{BerzakRK15}.
These studies are instrumental for language teaching and
student evaluation \citep{smith2001learner}, and can
improve NLP applications
such as authorship profiling \citep{estival2007author} or grammatical
error correction \citep{chodorow2010utility}.
Most of these studies utilize
techniques that are motivated by the same abstract principles
associated with L1 influences on the target language.

To the best of our knowledge, our work is the first to address both
translations and non-native language under a unifying computational framework, and
in particular to compare both with native language.

\section{Methodology and experimental setup}
\label{sec:exp-setup-class}

\subsection{Dataset}
Our dataset\footnote{The dataset is available at
\url{http://nlp.unibuc.ro/resources.html}}
is based on the highly homogeneous corpus of the European Parliament Proceedings
\citep{Koehn05Europarl}. Note that the proceedings are produced as follows:
\begin{inparaenum}[(1)]
\item the utterances of the speakers are transcribed;
\item the transcriptions are sent to the speaker who may suggest
  minimal editing without changing the content;
\item the edited version is then translated by native speakers. Note
  in particular that the texts are \emph{not} a product of
  simultaneous interpretation.
\end{inparaenum}

In this work we utilize a subset of Europarl in which each sentence is
manually annotated with speaker information, including
the EU state represented and the original language in which the sentence was uttered \citep{nnt-corpus}.
The texts in the corpus are uniform in terms of style, respecting the
European Parliament's formal standards. Translations are produced by native English speakers and all
non-native utterances are selected from members not representing UK or Ireland.
Europarl N consists of texts delivered by native speakers from England.

Table \ref{tbl:corpus-stats} depicts statistics of the dataset.%
\footnote{Appendix A provides details on the distribution of NN and T texts by various L1s.}
In contrast to other learner corpora such as ICLE \citep{icle}, EFCAMDAT \citep{efcamdat} or TOEFL-11 \citep{toefl}, this corpus
contains translations, native, and non-native English of high proficiency speakers.
Members of the European Parliament have the right to use any of the EU's 24 official
languages when speaking in Parliament, and the fact that some of them
prefer to use English suggests a high degree of
confidence in their language skills.

\begin{table}[hbt]
\centering
\resizebox{\columnwidth}{!}{
\begin{tabular}{l|rrr}
\textbf{sub-corpus} & \textbf{sentences} & \textbf{tokens} & \textbf{types} \\ \hline
native (N)  & 60,182 & 1,589,215 & 28,004 \\
non-native (NN) & 29,734 & 783,742 & 18,419 \\
translated (T) & 738,597 & 22,309,296 & 71,144 \\ \hline
total & 828,513 & 24,682,253 & 117,567
\end{tabular}
}%
\caption{Europarl corpus statistics: native, non-native and translated texts.}
\label{tbl:corpus-stats}
\end{table}

\subsection{Preprocessing}
All datasets were split by sentence, cleaned (text lowercased,
punctuation and empty lines removed) and tokenized using the Stanford
tools \citep{manning-EtAl:2014:P14-5}. For the classification
experiments we randomly shuffled the sentences within each language variety to
prevent interference of other artifacts (e.g., authorship, topic) into the
classification procedure. We divided the data into chunks of
approximately 2,000 tokens, respecting sentence boundaries, and
normalized the values of lexical features by the number of tokens in
each chunk. For classification we used Platt's sequential minimal
optimization algorithm \citep{Keerthi2001,weka} to train support vector
machine classifiers with the default linear kernel.

In all the experiments we used (the maximal) equal amount of data from
each category, thus we always randomly down-sampled the datasets in
order to have a comparable number of examples in each class;
specifically, 354 chunks were used for each language variety: N, NN and T.

\subsection{Features}
\label{sec:features}
The first feature set we utilized for the classification tasks
comprises \emph{function words} (FW), probably the most popular choice
ever since \citet{mostWallace} used it successfully for the Federalist
Papers. Function words proved to be suitable features for multiple
reasons:%
\begin{inparaenum}[(1)]
\item they abstract away from contents and are therefore less biased by topic;
\item their frequency is so high that by and large they are assumed to
  be selected unconsciously by authors;
\item although not easily interpretable, they are assumed to reflect
  grammar, and therefore facilitate the study of how structures are
  carried over from one language to another.
\end{inparaenum}
We used the list of approximately 400 function words provided in
\citet{koppel-ordan:2011:ACL-HLT2011}.

A more informative way to capture (admittedly shallow) syntax is to
use \emph{part-of-speech (POS) trigrams}. Triplets such as PP
(personal pronoun) + VHZ (\textnl{have}, 3sg present) + VBN
(\textnl{be}, past participle) reflect a complex tense form,
represented distinctively across languages. In Europarl, for example,
this triplet is highly frequent in translations from Finnish and
Danish and much rarer in translations from Portuguese and Greek. In
this work we used the top-3,000 most frequent POS trigrams in each
corpus.

We also used \emph{positional token frequency}
\citep{Grieve:2007}. The feature is defined as counts of words
occupying the first, second, third, penultimate and last positions in
a sentence. The motivation behind this feature is that sentences open
and close differently across languages, and it should be expected that
these opening and closing devices will be transferred from \lone\ if
they do not violate the grammaticality of the target
language. Positional tokens were previously used for translationese
identification \citep{vered:noam:shuly} and for native language
detection \citep{nisioi2015a}.

Translations are assumed to exhibit \emph{explicitation}: the tendency
to render implicit utterances in the source text more explicit in the
translation product. For example, causality, even though not always
explicitly expressed in the source, is expressed in the target by the
introduction of cohesive markers such as \textnl{because}, \textnl{due
  to}, etc. \citep{Blum-Kulka:1986}. Similarly,
\citet{hinkel2001matters} conducted a comparative analysis of
\emph{explicit cohesive devices} in academic texts by non-native
English students, and found that cohesive markers are distributed
differently in non-native English productions, compared to their
native counterparts. To study this phenomenon, we used the set of over
100 cohesive markers introduced in \citet{hinkel2001matters}.

\section{The status of constrained language}
\label{sec:classification}
To establish the unique nature of each language variety in our dataset,
we perform multiple pairwise binary classifications between N, NN,
and T, as well as three-way classifications.
Table~\ref{tbl:classification} reports the results; the
figures reflect average ten-fold cross-validation accuracy (the best
result in each column is boldfaced).

In line with previous works 
(see Section~\ref{sec:related-work}), classification of N--T, as well
as N--NN, yields excellent results with most features and feature
combinations. NN--T appears to be easily
distinguishable as well; specifically, FW+POS-trigrams combination
with/without positional tokens yields 99.57\% accuracy. The word
\textnl{maybe} is among the most discriminative feature for NN vs.\ T,
being overused in NN, as opposed to \textnl{perhaps}, which exhibits a
much higher frequency in T; this may indicate a certain degree of
formality, typical of translated texts \citep{Olohan:2003}. The words
\textnl{or}, \textnl{which} and \textnl{too} are considerably more
frequent in T, implying higher sentence complexity. This trait is also
reflected by shorter NN sentences, compared to T: the average sentence
length in Europarl is 26 tokens for NN vs.\ 30 for T. Certain
decisiveness devices (\textnl{sure}, \textnl{very}) are underused in
T, in accordance with \cite{Toury:1995}'s law of standardization
\citep{vanderauwera1985dutch}. The three-way classification yields
excellent results as well; the highest accuracy is obtained using
FW+positional tokens with/without POS-trigrams.

\begin{table}[hbt]
\centering
\resizebox{\columnwidth}{!}{
\begin{tabular}{@{}l|ccc|c}
feature / dataset   & N-NN & N-T & NN-T & 3-way\\
\hline
FW                  & 98.72 & 98.72 & 96.89 & 96.60 \\
POS (trigrams)      & 97.45 & 98.02 & 97.45 & 95.10 \\
pos. tok            & 99.01 & 99.01 & 98.30 & 98.11 \\
cohesive markers    & 85.59 & 87.14 & 82.06 & 74.19 \\
\hline
FW+POS              & 99.43 & 99.57 & \textbf{99.57} & 99.34 \\
FW+pos. tok         & 99.71 & 99.85 & 98.30 & \textbf{99.52} \\
POS+pos. tok        & 99.57 & 99.57 & 99.01 & 99.15 \\
FW+POS+pos. tok     & \textbf{99.85} & \textbf{99.85} & \textbf{99.57} & \textbf{99.52} \\
\end{tabular}
}
\caption{Pairwise and three-way classification results of N, NN and T texts.}
\label{tbl:classification}
\end{table}

A careful inspection of the results in Table~\ref{tbl:classification}
reveals that NN--T classification is a slightly yet systematically
harder task than N--T or N--NN; this implies that NN and T texts are
more similar to each other than either of them is to N.

To emphasize this last point, we analyze the separability of the three
language varieties by applying unsupervised classification. We perform \emph{bisecting KMeans}
clustering procedure previously used for unsupervised identification of translationese by \citet{TACL618}.
Clustering of N, NN and T using function words into three clusters
yields high accuracy, above 90\%. For the sake of clusters' visualization in a bidimensional plane,
we applied principal component analysis for dimensionality reduction.

\begin{figure}[hbt]
\vspace*{0.27em}
\centering
\begin{tabular}{cc}
\hspace{-0.75em}
\includegraphics[width=3.85cm]{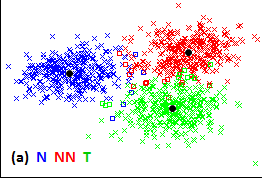} &
 \hspace{-1.25em}
\includegraphics[width=3.85cm]{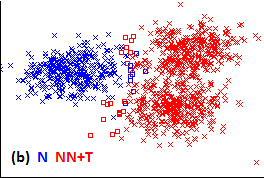}
\end{tabular}
\caption{Clustering of N, NN and T into three (a) and two (b) clusters using function words. Clusters'
centroids in (a) are marked by black circles; square sign stands for instances clustered wrongly.}
\label{fig:clustering}
\end{figure}

The results are depicted in Figure~\ref{fig:clustering}~(a). Evidently, NN and T exhibit higher
mutual proximity than either of them with N. Fixing the number of expected clusters
to~2 further highlights this observation, as demonstrated in
Figure~\ref{fig:clustering}~(b): both NN and T instances were assigned
to a single cluster, distinctively separable from the N cluster.

We conclude that the three language varieties (N, NN, and T) constitute three
different, distinguishable ontological categories, characterized by
various lexical, syntactic and grammatical properties; in particular,
the two varieties of constrained language (NN and T) represent two
distinct linguistic systems.
Nevertheless, we anticipate NN and T to share more common tendencies and regularities,
when compared to N.
In the following sections, we put this hypothesis to the test.

\section{\lone-independent similarities}
\label{sec:similarities}

In this section we address \lone-independent similarities between NN
and T, distinguishing them from N. We focus on characteristics which
are theoretically motivated by translation studies and which are
considered to be \lone-independent, i.e., unrelated to cross-linguistic influences.
We hypothesize that linguistic devices over- or under-represented in
translation would behave similarly in highly competent non-native
productions, compared to native texts.

To test this hypothesis, we realized various linguistic phenomena as
properties that can be easily computed from N, NN and T texts. We
refer to the computed characteristics as \emph{metrics}.  Our
hypothesis is that NN metric values will be similar to T, and that
both will differ from N. We used equally-sized texts of 780K tokens
for N, NN and T; the exact computation is specified for each metric.

\begin{figure*}[hbt]
\centering
\begin{tikzpicture}[font=\small]
  \centering
  \begin{axis}[
        ybar, axis on top,
        height=4.75cm, width=15.5cm,
        bar width=0.6cm,
        yminorgrids, tick align=inside,
        ymin=0.290, ymax=0.400,
        axis x line*=bottom,
        axis y line*=right,
        y axis line style={opacity=0},
        tickwidth=0pt,
        enlarge x limits=true,
        legend style={
            at={(0.5,-0.2)},
            anchor=north,
            legend columns=-1,
            /tikz/every even column/.append style={column sep=0.5cm}
        },
        ylabel={zero-one scale normalized value},
        xticklabels={lexical richness$^*$, mean word rank$^*$, pronouns$^*$, collocation (types)$^*$,
        transitions},
        xticklabel style={anchor=base,yshift=-\baselineskip},
        xtick=data,
    ]
    \addplot [draw=none, fill=black!20] coordinates {
    (0,0.356)(1,0.334)(2,0.393)(3,0.382)(4,0.303)};
    \addplot [draw=none, fill=black!80] coordinates {
    (0,0.332)(1,0.332)(2,0.310)(3,0.311)(4,0.377)};
    \addplot [draw=none, fill=black!50] coordinates {
    (0,0.312)(1,0.332)(2,0.296)(3,0.307)(4,0.320)};
     \legend{native (N), translated (T), non-native (NN)}
  \end{axis}
  \end{tikzpicture}
\label{fig:universal-metrics}
\caption{Metric values in N, NN and T. Tree-way differences are significant in all metric categories and
``*'' indicates metrics with higher pairwise similarity of NN and T, compared individually to N.}
\label{fig:universal}
\end{figure*}
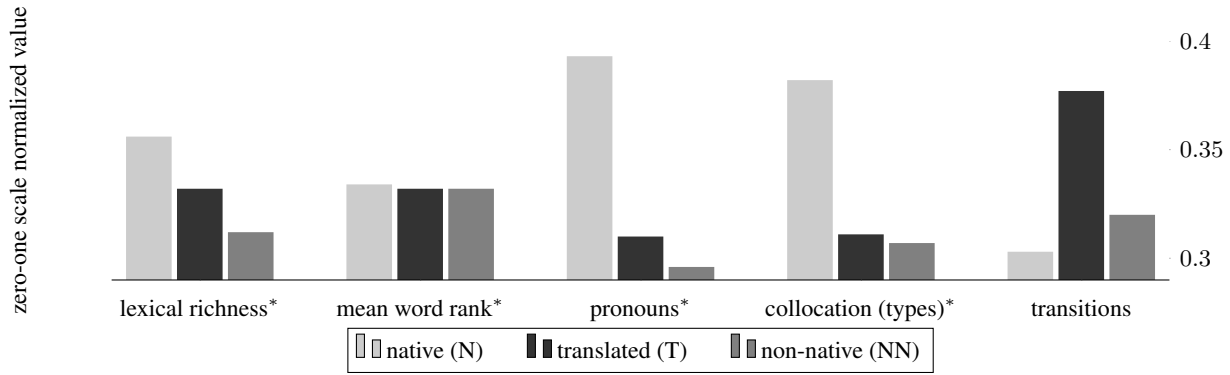

For the sake of visualization, the three values of each metric (for N, NN and T)
were zero-one scaled by total-sum
normalization. Figure~\ref{fig:universal} graphically depicts the
normalized metric values. We now describe and motivate each metric.
We analyze the results in Section~\ref{sec:universal-analysis} and
establish their statistical significance in
Section~\ref{sec:significance}.

\paragraph{Lexical richness}
Translated texts tend to exhibit less lexical diversity
\citep{Al-Shabab:1996}. \citet{Blum-Kulka:1986} suggested that
translated texts \textit{make do with less words}, which is reflected by
their lower type-to-token ratio (TTR) compared to that of native
productions. We computed the TTR metric by dividing the number of
unique (lemmatized) tokens by the total number of tokens.

\paragraph{Mean word rank}
\citet{Halverson:2003} claims that translators use more prototypical
language, i.e., \textit{they regress to the mean}
\citep{Shlesinger:1989}. We, therefore, hypothesize that rarer words are
used more often in native texts than in non-native productions and
translationese. To compute this metric we used a BNC-based ranked list
of 50K English words\footnote{\url{https://www.kilgarriff.co.uk} we used the list extracted from both
spoken and written text.},
excluding the list of function words (see
Section~\ref{sec:features}). The metric value was calculated by
averaging the rank of all tokens in a text; tokens that do not appear
in the list of 50K were excluded.

%

\paragraph{Collocations}
Collocations are distributed differently in translations and in
originals \citep{Toury:1980,Kenny:2001}. Common and frequent
collocations are used almost subconsciously by native speakers, but
will be subjected to a more careful choice by translators and,
presumably, by fluent non-native speakers
\citep{erman2014nativelike}. For example, the phrase \textnl{make
  sure} appears twice more often in native Europarl texts than in NN,
and five times more than in T; \textnl{bear in mind} has almost double
frequency in N, compared to NN and T. Expressions such as: \textnl{bring
  forward}, \textnl{figure out}, \textnl{in light of}, \textnl{food
  chain} and \textnl{red tape} appear dozens of times in N, as opposed
to zero occurrences in NN and T Europarl texts. This metric is defined
by computing the frequency of idiomatic expressions\footnote{Idioms
  were taken from
  \url{https://en.wiktionary.org/wiki/Category:English_idioms}. The
  list was minimally cleaned up.} in terms of types.

\paragraph{Cohesive markers}
Translations were proven to employ cohesion intensively
\citep{Blum-Kulka:1986,Overas:1990,koppel-ordan:2011:ACL-HLT2011}. Non-native
texts tend to use cohesive markers differently as well:
\emph{sentence transitions}, the major cohesion
category, was shown to be overused by non-native speakers regardless
of their native language \citep{hinkel2001matters}. The metric is
defined as the frequency of sentence transitions in the three
language varieties.

Qualitative comparison of various markers between NN and T
productions, compared to N in the Europarl texts, highlights this
phenomenon: \textnl{in addition} is twice as frequent in NN and T than
in N; \textnl{according}, \textnl{at the same time} and \textnl{thus}
occur three times more frequently in NN and T, compared to N;
\textnl{moreover} is used four times more frequently; and \textnl{to
  conclude} is almost six times more frequent.

\paragraph{Personal pronouns}
We expect both non-native speakers and translators to spell out
entities (both nouns and proper nouns) more frequently, as a means of
\emph{explicitation} \citep{olohan2002leave}, thus leading to
under-use of personal pronouns, in contrast to native texts.
As an example, \textnl{his} and \textnl{she} are twice more frequent in N than in NN and T.

We define this metric as the frequency of (all) personal and possessive pronouns used in the three language varieties.
The over-use of personal pronouns in N utterances, is indeed balanced out by lower frequency
of proper and regular nouns in these texts, compared to T and NN.\footnote{Normalized frequencies of nouns and proper nouns are
0.323, 0.331 and 0.345 for N, T, and NN, respectively.}

\subsection{Analysis}
\label{sec:universal-analysis}
Evidently (see Figure~\ref{fig:universal}), translationese and
non-native productions exhibit a consistent pattern in both datasets,
compared to native texts: NN and T systematically demonstrate lower
metric values than N for all characteristics (except sentence
transitions, where both NN and T expectedly share a higher value).
All metrics except mean word rank exhibit substantial (sometimes dramatic) differences between N, on the one hand,
and NN and T, on the other, thus corroborating our hypothesis.
Mean word rank exhibits a more moderate variability in the
three language varieties, yielding near identical value in NN and T;
yet, it shows excessive usage in N.

The differences between metric values are statistically significant
for all metrics (Section~\ref{sec:significance}). Moreover, in all
cases (except transitions), the difference between NN and T metrics is
significantly lower than the difference between either of them and N,
implying a higher proximity of NN and T distributions, compared
individually to N. This finding further emphasizes the common
tendencies between NN and T.

As shown in Figure~\ref{fig:universal}, NN and T are systematically
and significantly different from N. Additionally, we can see that T is
consistently positioned between N and NN (except for sentence transitions),
implying that translations produced
by native speakers tend to resemble native utterances to a higher degree
than non-native productions. 

\subsection{Statistical significance}
\label{sec:significance}
Inspired by the results depicted in Figure~\ref{fig:universal}, we now put to test two statistical hypotheses:
\begin{inparaenum}[(1)]
\item N, NN and T productions do not represent identical underlying distributions, i.e., at least one  pair is distributed differently; and consequently,
\item NN and T productions exhibit higher similarity (in terms of \emph{distance}) than either of them with N.
\end{inparaenum}
We test these hypotheses by applying the \emph{bootstrapping} statistical analysis.

Bootstrapping is a statistical technique involving random re-sampling
(with replacement) from the original sample; it is often used to
assign a measure of accuracy (e.g., a confidence interval) to an
estimate. Specifically, let $\native$, $\nonnative$ and $\translated$
denote native, non-native and translated sub-corpora of equal size
(780K tokens). Let $\corpusall$ denote the concatenation of all three
sub-corpora, resulting in a total of 2,340M tokens. We further denote
a function computing a metric $m$ by $\fmetric$; when applied to
$\corpus$, its value is $\fmetriccor$.  The sum of pairwise distances
between the three individual dataset metrics is denoted by
$\distotal$:

\setlength\abovedisplayskip{0pt}
\begin{equation*}
\begin{split}
&\distotal = |\fmetric(\native)-\fmetric(\nonnative)| + \\
&|\fmetric(\native)-\fmetric(\translated)| + |\fmetric(\nonnative)-\fmetric(\translated)|
\end{split}
\end{equation*}
\setlength\abovedisplayskip{0pt}

High values of $\distotal$ indicate a difference between the three
language varieties. To examine whether the observed $\distotal$ is high beyond
chance level, we use the bootstrap approach, and repeat the following
process 1,000 times:\footnote{This sample size is proven sufficient by
  the highly significant results (very low \textit{p}-value).} we
sample $\corpusall$ with replacement (at sentence granularity),
generating in the j-th iteration equal-sized samples
$\widehat\native^{j}$, $\widehat\nonnative^{j}$,
$\widehat\translated^{j}$. The corresponding distance estimate,
therefore, is:

\setlength\abovedisplayskip{0pt}
\begin{equation*}
\begin{split}
&\widehat\distotal^{j} = |\fmetric(\widehat\native^{j})-\fmetric(\widehat\nonnative^{j})| + \\
&|\fmetric(\widehat\native^{j})-\fmetric(\widehat\translated^{j})| + |\fmetric(\widehat\nonnative^{j})-\fmetric(\widehat\translated^{j})|
\end{split}
\end{equation*}
\setlength\abovedisplayskip{0pt}

We repeat random re-sampling and computation of
$\widehat\distotal^{j}$ 1,000 times, and estimate the \textit{p}-value
of $\widehat\distotal$ by calculation of its percentile within the
series of (sorted) $\widehat\distotal^{j}$ values, where
$j \in (1,\ldots,1000)$. In all our experiments the original distance
$\distotal$ exceeds the maximum estimate in the series of
$\widehat\distotal^{j}$, implying highly significant difference, with
\textit{p}-value\textless$0.001$ for all metrics.

In order to stress this outcome even further, we now test whether (the
constrained) NN and T exhibit higher pairwise similarity, as opposed
to N. We achieve this by assessment of the distance between NN and T
productions, compared to the distance between N and its closest
production (again, in terms of distance): either NN or T. We sample
$\native$, $\nonnative$ and $\translated$ (with replacement)
separately, constructing $\widetilde\native$, $\widetilde\nonnative$
and $\widetilde\translated$, respectively, and define the following
distance function:

\setlength\abovedisplayskip{0pt}
\begin{equation*}
\widetilde\disnntn^{j}\!=\!|\fmetric(\widetilde\native^{j})\!-\!\fmetric(\widetilde\corpus_{K}^{j})|\!-
\!|\fmetric(\widetilde\nonnative^{j})\!-\!\fmetric(\widetilde\translated^{j})|
\end{equation*}
\setlength\abovedisplayskip{0pt}

\noindent where
\setlength\abovedisplayskip{0pt}
\begin{equation*}
K\!=\!
\begin{cases}
\mbox{NN} &\mbox{if}\;|\fmetric(\native)-\fmetric(\nonnative)|< \\
& |\fmetric(\native)-\fmetric(\translated)| \\
\mbox{T} &\mbox{otherwise}
\end{cases}
\end{equation*}
\setlength\abovedisplayskip{0pt}

We repeat re-sampling and computation of $\widetilde\disnntn^{j}$
1,000 times for each metric value in both datasets and sort the
results. The end points of the 95\% confidence interval are defined by
estimate values with 2.5\% deviation from the minimum
(\emph{min-end-point}) and the maximum (\emph{max-end-point})
estimates. We assess the \textit{p}-value of the test by inspecting
the estimate underlying the min-end-point; specifically, in case the
min-end-point is greater than $0$, we consider
\textit{p}\textless$0.05$. Metric categories exhibiting higher NN-T
similarity than either N-NN or N-T are marked with ``*'' in
Figure~\ref{fig:universal}.

\section{\lone-related similarities}
\label{sec:interference}
We hypothesize that both varieties of constrained language exhibit similar
(lexical, grammatical, and structural) patterns due to the influence of
\lone\ over the target language. Consequently, we anticipate that
non-native productions of speakers of a certain native language
(\lone) will be closer to translations from \lone\ than to
translations from other languages.

Limited by the amount of text available for each individual language,
we set out to test this hypothesis by inspection of two language
\emph{families}, Germanic and Romance. Specifically, the Germanic
family consists of NN texts delivered by speakers from Austria,
Germany, Netherlands and Sweden; and the Romance family includes NN
speakers from Portugal, Italy, Spain, France and Romania. The
respective T families comprise translations from Germanic and Romance
originals, corresponding to the same
countries. Table~\ref{tbl:families-stats} provides details on the
datasets.

\begin{table}[hbt]
\centering
\resizebox{\columnwidth}{!}{%
\begin{tabular}{l|rrr}
 & \textbf{sentences} & \textbf{tokens} & \textbf{types} \\ \hline
Germanic NN & 5,384 & 132,880 & 7,841 \\
Germanic T & 269,222 & 7,145,930 & 43,931 \\ \hline
Romance NN & 6,384 & 180,416 & 9,838 \\
Romance T & 307,296 & 9,846,215 & 49,925 \\
\end{tabular}
}
\caption{Europarl Germanic and Romance families: NN and T.}
\label{tbl:families-stats}
\end{table}

We estimate \lone-related traces in the two varieties of constrained
language by the fitness of a translationese-based \emph{language
  model} (LM) to utterances of non-native speakers from the same
language family. Attempting to trace structural and grammatical,
rather than content similarities, we compile five-gram \emph{POS}
language models from Germanic and Romance translationese (GerT and
RomT, respectively).\footnote{For building LMs we used the closed vocabulary of Penn Treebank POS tag set.}
We examine the prediction power of these models
on non-native productions of speakers with Germanic and Romance native
languages (GerNN and RomNN), hypothesizing that an LM compiled from
Germanic translationese will better predict non-native productions of
a Germanic speaker and vice versa. The fitness of a language model to
a set of sentences is estimated in terms of \emph{perplexity}
\citep{JelMerBahBak}.

For building and estimating language models we used the KenLM toolkit
\citep{heafield2011kenlm}, employing modified Kneser-Ney smoothing
without pruning. Compilation of language-family-specific models was
done using 7M tokens of Germanic and Romance translationese each; the
test data consisted of 5350 sentences of Germanic and Romance
non-native productions. Consequently, for perplexity experiments with
individual languages we utilized 500 sentences from each language. We
excluded OOVs from all perplexity computations.

Table~\ref{tbl:ppl-family} reports the results. Prediction of GerNN by
the GerT language model yields a slightly lower perplexity (i.e., a
better prediction) than prediction by RomT. Similarly, RomNN is much
better predicted by RomT than by GerT. These differences are
statistically significant: we divided the NN texts into 50 chunks of
100 sentences each, and computed perplexity values by the two LMs for
each chunk. Significance was then computed by a two-tailed paired
t-test, yielding p-values of 0.015 for GerNN and 6e-22 for RomNN.

\begin{table}[hbt]
\centering
\resizebox{\columnwidth}{!}{%
\begin{tabular}{l|cccl|c}
LM / NN & GerNN & & & LM / NN & RomNN \\
\hline
GerT    & 8.77 & & & GerT  & 8.64 \\
RomT    & 8.79 & & & RomT  & 8.43 \\
\end{tabular} %
}
\caption{Perplexity: fitness of Germanic and Romance translationese LMs to Germanic and Romance NN test sets.}
\label{tbl:ppl-family}
\end{table}

\begin{figure}[hbt]
\centering
\hspace{-1em}
\begin{tikzpicture}
\begin{axis}[
    axis lines=middle,
    xlabel=Perplexity by GerT,
    ylabel=Perplexity by RomT,
    ylabel near ticks,
    xlabel near ticks,
    x label style={font=\footnotesize},
    y label style={font=\footnotesize},
    xmin=11.4, xmax=12.85,
    ymin=11.4, ymax=12.85,
    xtick=\empty, ytick=\empty
]
\addplot [
    mark=*,
    only marks,
    nodes near coords,
    mark options={fill=white},
    point meta=explicit symbolic,
    visualization depends on=\thisrow{alignment} \as \alignment,
    every node near coord/.style={anchor=\alignment,font=\footnotesize}
    ] table [meta index=2] {
x           y           label          alignment
11.6746	    11.9368     Austria         -90
12.2198	    12.3045     Netherlands     -90
11.8418	    11.8993     Sweden          50
11.9694	    11.9351     Germany         -90
};
\addplot [
    mark=*,
    only marks,
    nodes near coords,
    mark options={fill=black},
    point meta=explicit symbolic,
    visualization depends on=\thisrow{alignment} \as \alignment,
    every node near coord/.style={anchor=\alignment,font=\footnotesize}
    ] table [meta index=2] {
x           y           label          alignment
12.1342	    12.0846     France          -180
11.9573	    11.8859     Italy           -180
11.7082	    11.5707     Romania         -180
12.8399	    12.3847     Spain           50
11.5786	    11.4424     Portugal        -180
};
\addplot [domain=11.4:12.75,dashed] {x};
\end{axis}
\end{tikzpicture}
\vspace{-0.5em}
\caption{Perplexity of the GerT and RomT language models with respect to non-native utterances of speakers from various countries.}
\label{fig:ppl-language}
\end{figure}
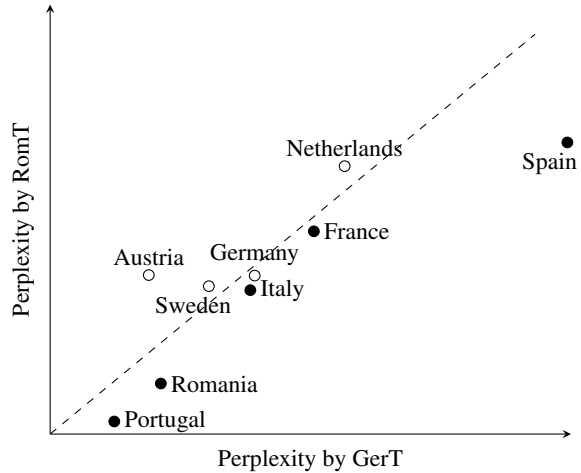

As a further corroboration of the above result, we computed the
perplexity of the GerT and RomT language models with respect to the
language of NN speakers, this time distinguishing speakers by their
country of origin. We used the same language models and non-native
test chunks of 500 sentences each. Inspired by the outcome of the
previous experiment, we expect that NN productions by Germanic
speakers will be better predicted by GerT LM, and vice
versa. Figure~\ref{fig:ppl-language} presents a scatter plot with the
results.

A clear pattern, evident from the plot, reveals that all English texts
with underlying Romance native languages (under the diagonal) are
better predicted (i.e., obtain lower perplexity) by the RomT LM.
All Germanic native languages (except German), on the other
hand, are better predicted by the GerT LM. This finding
further supports the hypothesis that non-native productions and
translationese tend to exhibit similar \lone-related traits.

\section{Conclusion}
\label{sec:conclusion}
We presented a unified computational approach for studying constrained
language, where many of the features were theoretically motivated. We
demonstrated that while translations and non-native productions are two
distinct language varieties, they share similarities that stem from lower
lexical richness, more careful choice of idiomatic expressions and
pronouns, and (presumably) subconscious excessive usage of
explicitation cohesive devices. More dramatically, the language
modeling experiments reveal salient ties between the native language
of non-native speakers and the source language of translationese,
highlighting the unified \lone-related traces of \lone\ in both
scenarios.  Our findings are intriguing: native speakers and
translators, in contrast to non-native speakers, use their native
language, yet translation seems to gravitate towards non-native
language use.

The main contribution of this work is empirical, establishing the
connection between these types of language production. While we
believe that these common tendencies are not incidental, more research
is needed in order to establish a theoretical explanation for the
empirical findings, presumably (at least partially) on the basis of
the cognitive load resulting from the simultaneous presence of two
linguistic systems. We are interested in expanding the
preliminary results of this work: we intend to replicate the
experiments with more languages and more domains, investigate
additional varieties of constrained language and employ more complex
lexical, syntactic and discourse features. We also plan to investigate
how the results vary when limited to specific \lone s.

\section*{Acknowledgments}
This research was supported by the Israeli Ministry of Science and Technology.
We are immensely grateful to Yuval Nov for much advice and helpful suggestions.
We are also indebted to Roy Bar-Haim, Anca Bucur, Liviu P. Dinu, and Yosi Mass for their support and guidance
during the early stages of this work, and we thank our anonymous reviewers for their valuable insights.

\bibliographystyle{acl_natbib}
\bibliography{all}

\newpage
\section*{Appendix A - Distribution of L1s in Translations and Non-native Texts}
\label{appendix-dist}
We assume that native languages of non-native speakers are highly correlated with
(although not strictly identical to) their country of origin.
\begin{table}[h]
\centering
\begin{tabular}{@{}l|rr@{}}
country of origin & tokens(T) & tokens(NN) \\
\hline
Austria         & -         & 2K  \\
Belgium         & -         & 67K \\
Bulgaria        & 25K       & 6K  \\
Cyprus          & -         & 35K \\
Czech Republic  & 21K       & 3K  \\
Denmark         & 444K      & 14K \\
Estonia         & 32K       & 50K \\
Finland         & 500K      & 81K \\
France          & 3,486K    & 28K \\
Germany         & 3,768K    & 17K \\
Greece          & 944K      & 13K \\
Hungary         & 167K      & 38K \\
Italy           & 1,690K    & 15K \\
Latvia          & 38K       & 13K \\
Lithuania       & 177K      & 18K \\
Luxembourg      & -         & 46K \\
Malta           & 28K       & 40K \\
Netherlands     & 1,746K    & 64K \\
Poland          & 522K      & 36K \\
Portugal        & 1,633K    & 54K \\
Romania         & 244K      & 29K \\
Slovakia        & 88K       & 6K  \\
Slovenia        & 43K       & 1K  \\
Spain           & 1,836K    & 54K \\
Sweden          & 951K      & 52K \\

\end{tabular}
\caption{Distribution of L1s by country.}
\end{table}

\end{document}